\documentclass[runningheads]{llncs}
\usepackage[T1]{fontenc}
\usepackage{graphicx}
\usepackage{color,soul}
\usepackage{amssymb}
\usepackage{amsmath}
\usepackage{makecell}
\begin{document}
\title{Generalization Capabilities of Neural Cellular Automata for Medical Image Segmentation: A Robust and Lightweight Approach}
\titlerunning{Generalization Capabilities of Neural Cellular Automata}
%
\author{Steven~Korevaar\inst{1}\orcidID{0000-0002-6597-8523} \and
Ruwan~Tennakoon\inst{1}\orcidID{0000-0001-8909-5728} \and
Alireza~Bab-Hadiashar\inst{2}\orcidID{0000-0002-6192-2303}}
%
%
%
\authorrunning{Korevaar et al.}
%
\institute{School of  Computing Technologies, RMIT University, Melbourne, Australia.
\email{\{steven.korevaar, ruwan.tennakoon\}@rmit.edu.au}\\ \and
School of Engineering, RMIT University, Melbourne, Australia.\\
\email{abh@rmit.edu.au}}
%
%
%
\maketitle              
\begin{abstract}
In the field of medical imaging, the U-Net architecture, along with its variants, has established itself as a cornerstone for image segmentation tasks, particularly due to its strong performance when trained on limited datasets. Despite its impressive performance on identically distributed (in-domain) data, U-Nets exhibit a significant decline in performance when tested on data that deviates from the training distribution, out-of-distribution (out-of-domain) data. Current methodologies predominantly address this issue by employing generalization techniques that hinge on various forms of regularization, which have demonstrated moderate success in specific scenarios. This paper, however, ventures into uncharted territory by investigating the implications of utilizing models that are smaller by three orders of magnitude (i.e., x1000) compared to a conventional U-Net. A reduction of this size in U-net parameters typically adversely affects both in-domain and out-of-domain performance, possibly due to a significantly reduced receptive field. To circumvent this issue, we explore the concept of Neural Cellular Automata (NCA), which, despite its simpler model structure, can attain larger receptive fields through recursive processes. Experimental results on two distinct datasets reveal that NCA outperforms traditional methods in terms of generalization, while still maintaining a commendable IID performance.

\end{abstract}
\section{Introduction}
Identifying precise locations and boundaries of biological structures within medical imagery, i.e., segmentation is a crucial component that enables accurate diagnosis and treatment planning.
Fully convolutional encoder-decoder neural network architectures like U-Net \cite{Olaf_2015} have become the preferred method in image segmentation, specifically in medical imaging where the training data is scarce \cite{wang2022medical}. The success of these architectures can be attributed primarily to their ability to combine global information (via successive convolution and downsampling layers) and local information (through the use of skip connections between the encoder and decoder).

While U-Net-like architectures demonstrate impressive performance on test data that aligns with the same distribution as the training data (in-domain), they often fall short in achieving the required accuracy when the test data originates from a slightly different distribution (out-of-domain). Distribution shift is common in medical imaging, particularly due to the use of different reconstruction parameters, resolutions, or general scanning procedures by various hospitals or scanning machines. 
Current, methodologies for handling distribution shift (i.e., domain generalization methods), predominantly address this issue by employing techniques that hinge on various forms of regularization \cite{wang_generalizing_2022}, which have demonstrated moderate success in specific scenarios \cite{gulrajani2020search}, \cite{korevaar2023failure}.

In this paper, we ask the question ``\textit{Are large models, which are prone to overfitting on training domain, necessary for medical image segmentation?}''. As previously stated, models similar to U-Net incorporate convolution and down-sampling layers facilitating the capture of information about larger objects, thereby increasing the model size. 
Consequently, reducing the model size by eliminating layers (or stages) could lead to a significant decrease in performance. In contrast, this paper explores Neural Cellular Automata (NCA) architecture for image segmentation. An NCA has the potential to accumulate information from large areas within the image through the iterative application of a neural network (which can be a simple multi-layer perceptron) to a local neighborhood.
Recent work has shown that NCA achieves respectable performance in segmentation tasks while being more robust to certain data transformations: like cropping, scaling, and translation \cite{mednca_2023}. However, to the best of our knowledge, the generalization capabilities of NCA to naturally occurring distribution shifts remain unexplored.

In this paper, we evaluate the performance of standard U-Net and a Neural Cellular Automata (NCA), which has roughly three orders of magnitude (thousand times) fewer weights, on two real-world medical image segmentation tasks.
In both tasks, the U-Net marginally surpasses the NCA in terms of in-domain data performance. However, both models exhibit a significant performance drop when applied to out-of-domain data. In contrast, the NCA, despite achieving marginally lower in-domain accuracies, demonstrates a lesser performance drop when applied to out-of-domain data, resulting in better out-of-domain performance. 
Our observations present a compelling argument for the utilization of neural cellular automata when generalization is required for image segmentation tasks.

\subsection{Background}
\noindent \textbf{Cellular automata:} 
In computer science and mathematics, researchers such as Von Neumann, have been exploring a novel form of swarm intelligence: cellular automata \cite{von1966theory}. These automata which were designed to mimic very simple biological cells, operate on a graph life lattice of cells. Each cell can ``see'' the cells surrounding it, and based on the cells around it and a set of ``rules'' the cell will alter its own state. The most famous cellular automata is John Conway's ``Game of Life'' shown to the public in 1970, which has been shown to be Turing Complete. These cellular automata (CA) have been researched and used extensively in many fields, due to their simple design, which can lead to highly complex behavior.

In the case of medical imagery, CA has also seen some use for tasks like segmentation, classification, and image noise processing (noise removal and registration). However, these approaches have used hand-crafted cell update rules.
\cite{wongthanavasu2011cellular}, \cite{ION2023999}, \cite{6061452}, \cite{9544514}, \cite{SOMPONG2017231}, \cite{4381715}.
More recent work has shown the capacity for using deep learning techniques to learn the update rule-set, allowing for more widespread use on far more complex datasets and tasks. This process of learning the update has been called neural cellular automata, or cellular learning automata.
In practice, the update rule-set is most often implemented as a multi-layer perceptron.
In the field of medical imaging, it was shown that NCA demonstrates a surprising level of effectiveness when compared to far more complex and large deep learning techniques: particularly in their robustness to shifts in the input \cite{mednca_2023}. Given their demonstrated robustness to input shifts, one area comes to mind that must be explored: the problem of domain generalization.

\vspace{1em}
\noindent \textbf{Domain Generalization in Medical Imaging: }
The problem of domain generalization can be described as learning a model, $f_\theta : \mathcal{X} \longrightarrow y$, from $k$ differently distributed source datasets, $\left\{ S_k \sim P^{S^k}_{Xy}\right\}^K_{k=1}$, that can achieve low task error on a similar but different and unseen set of data: i.e., $S_T \sim P_{Xy}^{T}$, where $ P_{Xy}^{T} \neq  P_{Xy}^{S^K} ~\forall k \in [1\dots K]$. In medical imaging, the problem of domain shift robustness is significant due to the combination of small amounts of available training data, which limits a model's exposure to shifts, and to the variety and severity of domain shifts (different scanning machines, different scanning parameters on the same machine, and differences in patients from different hospitals). Both of these factors mean that typical deep learning methods often fail when tested on data from differing distributions (out-of-distribution, OOD). As such there has been extensive work in altering the learning procedures for traditional deep learning models to account for future potential domain shifts. Most methods revolve around regularizing the learning through additional loss functions, using additional data (self-supervised/unsupervised pre-training), or simulating additional data (data augmentation) \cite{wang_generalizing_2022} \cite{zhou_domain_2022}. However, given the success of the neural cellular automata's robustness, it may be more applicable to explore domain robustness using an architecture that natively lends itself well to generalization, as opposed to trying to retrofit generalization capacity into a model that does not support it natively. 


\section{Methodology}

In this work, we explore the generalisability of 2D cellular automata for medical image segmentation and compare it with standard U-Net. All architectures were assessed using the same training methodology on two medical imaging segmentation tasks: the Retinal Optical Coherence Tomography Fluid Challenge (RETOUCH) \cite{bogunovic_retouch_2019}, and the Multi-Disease, Multi-View \& Multi-Center segmentation challenge (MNM2) \cite{campello_multi-centre_2021}. 
The following subsections describe the architectures and the datasets used. 

\subsection{Network Architectures}

\noindent \textbf{Neural Cellular Automata:}
A 2D cellular automata can be defined as a combination of three components: (1) A matrix of ``cells'' of size $I \times J$, each with values $C_{i,j} \in \mathbb{R}^D$, (2) A neighborhood $\mathcal{N}(\cdot)$ that defines cell adjacency, and (3) A rule-set $\mathcal{R}(\cdot)$. At each iteration, $t$, the cells are updated using the rule-set, which is a function of the cell's current values, $C_{i,j}^t$, and the values of the cells in the neighborhood: 
\begin{equation}
     C_{i,j}^{t+1} = \mathcal{R}\left ( C_{i,j}^{t}, \left [  C_{p,q}^{t} \mid {\left\{ p, q\right\} \in \mathcal{N}(i,j)} \right ]\right )
\end{equation}
Contrary to typical cellular automata where the ruleset is manually designed, in this study, we employ a Neural Cellular Automata (NCA) where the ruleset, denoted as, $\mathcal{R}_\theta(\cdot)$, is parameterized by $\theta$ and represented as a neural network. While the aforementioned formulation bears a resemblance to Graph Neural Networks (GNN), there exists a significant distinction: in the context of NCA, the sequence of the neighborhood is preserved, whereas in GNN, the neighborhood is generally unordered.

\begin{figure}[!t]
    \centering
    \includegraphics[width=0.75\textwidth]{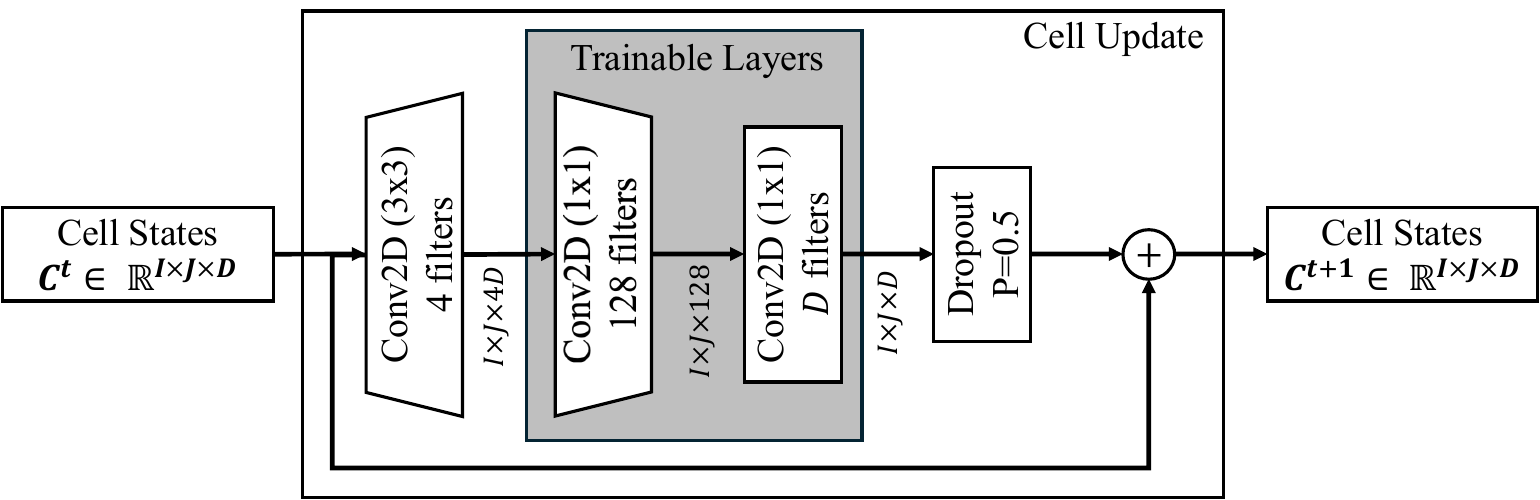}
    \caption{The Cell update model of the Neural Cellular Automata used in this work.}
    \label{fig:nca_model}
\end{figure}

In our implementation, we use a simple neighborhood of $3 \times 3$, and a cell state dimension $D = 32$. Each element in the cell state vector represents specific purposes: the first $d$ ($d=1$ for both datasets) elements are used to store the pixel values of the input image which remain static across the iterations, the next four (number of classes) are used for the output of the model which is run through a softmax function before computing the loss function, the remaining elements represent latent representation of each cell which is learned by the NCA.
The schematic representation of the neural network architecture of the rule-set, is provided in Figure \ref{fig:nca_model}. 
The first stage is aggregating the information from the neighborhood. For this we use static (non-trainable) 2D convolution layers with 4 filters: 1) An identity filter to send the cell's current information to the next step. 2) An average filter, which simply averages all the cells' states within the neighborhood. 3,4) Captures the Horizontal and vertical relationships in cells. The purpose of these fixed kernels is to reduce the dimensions of the input while providing adequate information for the cells to pass between each other. 
\begin{equation}
\theta^{[1]} =\left[
    \begin{bmatrix}
    0 & 0 & 0 \\
    0 & 1 & 0 \\
    0 & 0 & 0 
    \end{bmatrix} 
    \begin{bmatrix}
    \frac{1}{9} & \frac{1}{9} & \frac{1}{9} \\
    \frac{1}{9} & \frac{1}{9} & \frac{1}{9} \\
    \frac{1}{9} & \frac{1}{9} & \frac{1}{9} 
    \end{bmatrix} 
    \begin{bmatrix}
    -\frac{1}{8} & -\frac{2}{8} & -\frac{1}{8} \\
    0 & 0 & 0 \\
    \frac{1}{8} & \frac{2}{8} & \frac{1}{8} 
    \end{bmatrix}  
    \begin{bmatrix}
    -\frac{1}{8} & 0 & \frac{1}{8} \\
    -\frac{2}{8} & 0 & \frac{2}{8} \\
    -\frac{1}{8} & 0 & \frac{1}{8} 
    \end{bmatrix} 
    \right]
\end{equation}
The next two layers act as the cell update rule-set, the first of which is a linear layer that expands the $4\times D$ input to an arbitrary hidden layer length, $H$, with a $1 \times 1$ convolution layer followed by ReLU activation. In our implementation, we set $H=128$. The final layer is another $1 \times 1$ convolution layer, that reduces the cell states from $H$ back to the original cell state size, $C$. Overall, our NCA has $20,608$ trainable weights. 
To train the NCA we used DICE loss, which is calculated as the
proportion of the overlapping area of the predicted segmentation map and the
actual segmentation mask; it is expressed as follows:
\begin{equation}
    \mathcal{L}_{DICE} = \sum_{c} \sum_{\{i,j\} } \left( 1 - \frac{2 \cdot (\hat{y}_{i,j}^c \cap y_{i,j}^c)}{\hat{y}_{i,j}^c \cup y_{i,j}^c} \right) .
\end{equation}
Here, $\hat{y}_{i,j}^c$ represent the predicted softmax score for class $c$ at pixel $\{i,j\}$, and $y_{i,j}^c$ the ground truth.  

\vspace{.5em}
\noindent \textbf{Convolutional Neural Network (U-Net):}
The baseline architecture used in most medical image segmentation tasks is the residual U-Net. Which uses a sequence of downsizing and convolutional blocks to extract features from an image, and then another sequence that upsizes the features back to an output of the same size as the input: forming a ``U'' shape.

In our study, we employed two distinct versions of U-Net, each with a different amount of trainable parameters: the original U-Net \cite{Olaf_2015} and a more compact version, U-Net-Lite \cite{khan2022leveraging}. The original U-Net, being the more complex model, comprises four encoder stages and possesses approximately 15 million (14,789,700 to be precise) trainable parameters. This is nearly a thousand times greater than the parameter count of the NCA. Conversely, U-Net-Lite, which is designed with only three stages, has a significantly lower parameter count of 58,892. This is roughly triple the parameter count of the NCA, making it a more comparable alternative. The loss function is the same $\mathcal{L}_{DICE}$ used for NCA.

\vspace{.5em}
\noindent \textbf{Hyper-parameters and model selection} 
As mentioned earlier, these hyper-parameters of both models were chosen by observing the performance on an unseen subset of \textit{source training data} (without using any information from the target domain), as would occur in a practical situation. This methodology is key to fair comparison of generalization as outlined by Gulrajani \& Loper-Paz \cite{gulrajani2020search}. The number of cell update iterations varied: for training the number was uniformly random for every batch between 64 and 256 updates to provide some basic temporal regularization, and for evaluation, this was locked to 128 iterations. The models were trained for 100 epochs with a batch size of 32. The AdamW optimizer was used with a learning rate of 5e-4.


The hyper-parameters were chosen in the same way as for the NCA, by evaluating on an unseen set of source domain data. The optimization parameters remained constant: AdamW optimizer with a learning rate of 5e-4, trained for 100 epochs with a batch size of 32. 

\subsection{Datasets}
\begin{figure}[htp]
    \centering
    \includegraphics[width=0.65\textwidth]{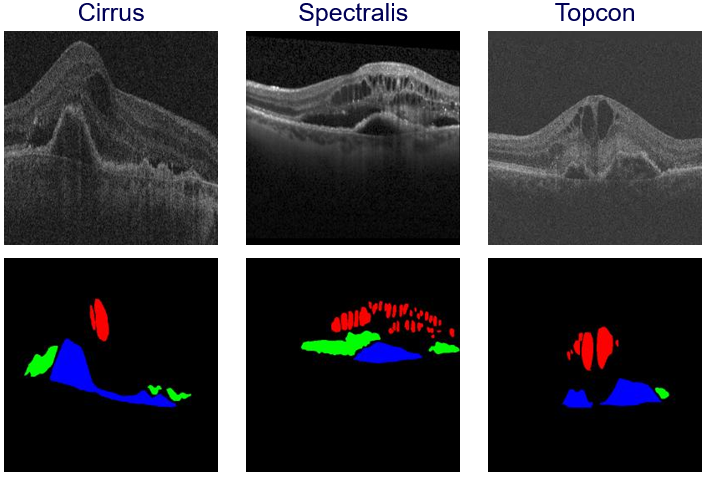}
    \caption{Sample images from the RETOUCH dataset: one input example (top row) and one segmentation map (bottom row) for each type of scanner in the dataset. Intraretinal fluid (red), Subretinal fluid (green), and Pigment Epithelial Detachment (blue).}
    \label{fig:retouch_samples}
\end{figure}

\noindent \textbf{Retouch:}
The Retinal Optical Coherence Tomography Fluid Challenge (Retouch) \cite{bogunovic_retouch_2019} involves segmenting three different forms of retinal fluids: Intraretinal fluid (red), Subretinal fluid (green), and Pigment Epithelial Detachment (blue). Each scan has ground truth segmentation masks that indicate the presence of the three retinal fluids. The dataset contains three subsets of data, with 24, 23, and 21 3D scans each with 100 slices approximately. Each domain consists of scans from different OCT machine vendors (Cirrus, Spectralis, and Topcon). Examples of the data can be seen in figure \ref{fig:retouch_samples}. 
The images were cropped to have a square aspect ratio around the main retinal layers in the image.

\vspace{.5em}
\noindent \textbf{MNM2 Segmentation Challenge:}
The Multi-Disease, Multi-View \& Multi-Center segmentation challenge (MNM2) \cite{campello_multi-centre_2021} \cite{10103611} is comprised of cardiac MRIs from three different scanner manufacturers: GE Medical Systems, Philips, and Siemens, with 53, 88, and 219 patients, respectively. The task is to segment different components of the heart: the left (LV) and right ventricle (RV) blood pools, and the left ventricular myocardium (MYO). The scans had values clipped between 0 and 2000 then normalized to a range of 0 to 1. As the challenge was also multi-view, only the end-systolic phase scans were used for training and testing \cite{campello_multi-centre_2021}.

\begin{figure}[htp]
    \centering
    \includegraphics[width=0.65\textwidth]{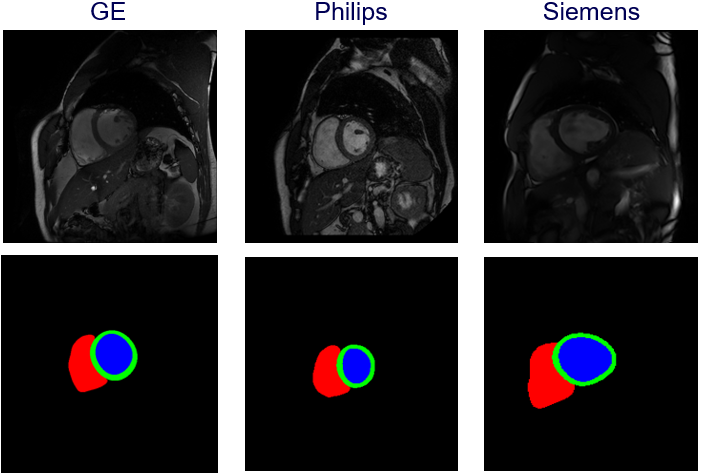}
    \caption{Sample images from the MnM2 dataset: one input example (top row) and one segmentation map (bottom row) for each type of scanner in the dataset. The left (LV) and right ventricle (RV) blood pools are blue and red, respectively, and the left ventricular myocardium (MYO) is green.}
    \label{fig:mnm_samples}
\end{figure}

\section{Results}
In our experimental setup; for each dataset, the model was trained on each pair of domains, while the third domain was reserved as an unseen target. This procedure was iterated three times, ensuring that each domain served as the target five times. The models were then validated on an unseen subset of data from the training domains, referred to as In-Distribution (IID), utilizing the DICE loss metric where the best model of the run was chosen for testing on the Out-Of-Distribution (OOD) domain. The final results are an average of five runs following the above strategy. The Dice segmentation accuracy for the RETOUCH and MNM2 datasets are documented in Tables \ref{tab:retouch_results} and \ref{tab:mnm_results}, respectively.

\begin{figure}[htp]
    \centering
    \includegraphics[width=0.95\textwidth]{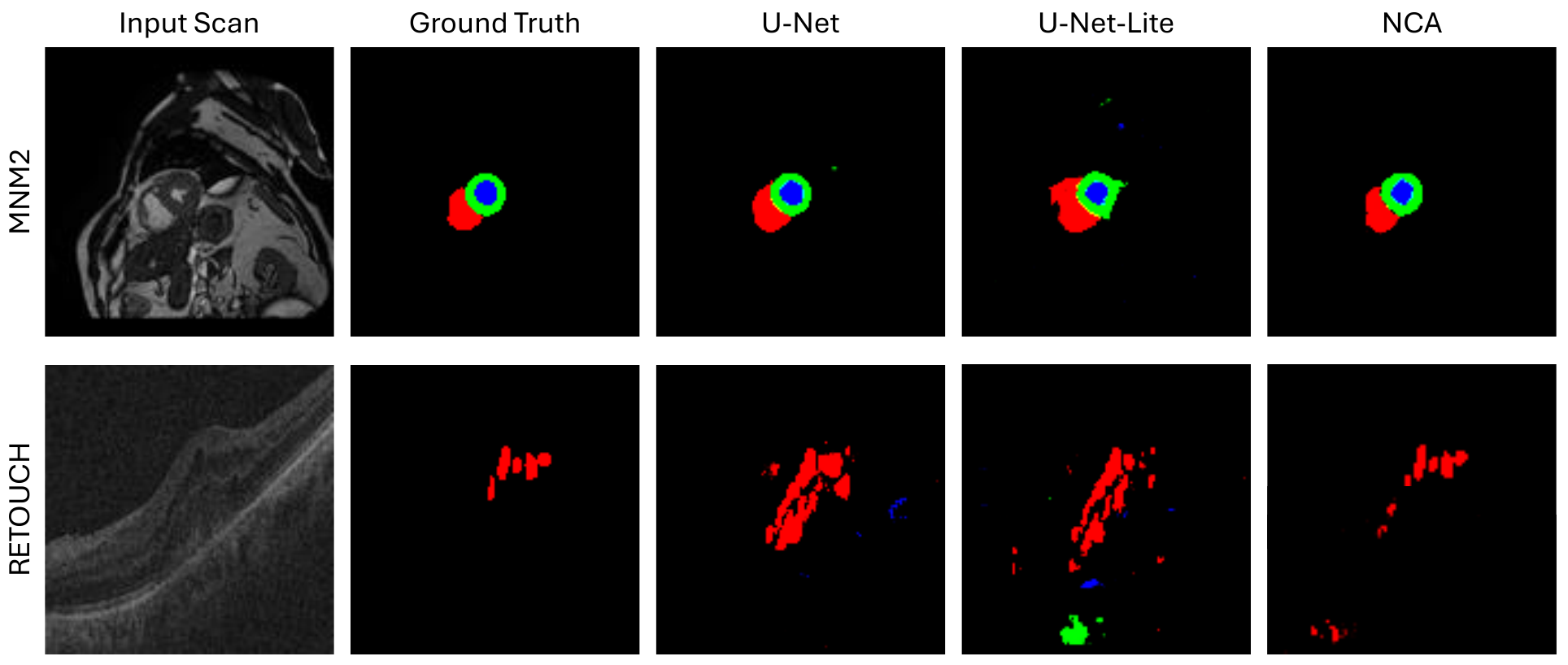}
    \caption{Randomly sample outputs from the out-of-distribution domain during testing. The MNM2 sample comes from the Philips scanner, and the RETOUCH sample from the Cirrus scanner. MNM2 labels: left and right ventricle blood pools are blue and red, and the left ventricular myocardium is green. RETOUCH labels: Intraretinal fluid is red, Subretinal fluid is green, and Pigment Epithelial Detachment is blue.}
    \label{fig:model_outputs}
\end{figure}

As can be seen in both tables \ref{tab:retouch_results} and \ref{tab:mnm_results} the neural cellular automata (NCA) has a higher average OOD Dice accuracy thann both U-Net versions, implying stronger generalization capacity. In the RETOUCH dataset the NCA outperforms the U-Nets in both OOD and IID accuracy, however given the better OOD performance, the NCA has a much lower generalisation gap. The shallow U-Net performs significantly worse in both generalization and in-distribution performance.
Some qualitative results across different models are shown in Figure \ref{fig:model_outputs}. The RETOUCH data also shows much higher generalisation gaps over MNM2, this can be visibly seen as the visual discrepancy between the different OCT scanners in the RETOUCH dataset is much more apparent over the differences between the MRI scanners in the MNM2 dataset.

\begin{table}[ht]
    \centering
    \caption{Average out-of-distribution Dice Accuracy on each domain for the RETOUCH segmentation dataset. Also includes the average performance on unseen validation (IID) data and the difference between IID and OOD accuracies.}
    \setlength{\tabcolsep}{6pt}
    \begin{tabular}{r|ccc|ccc}
    \hline

     &  &  &  & {Mean} & {Mean} & {Mean} \\
    
     & {Cirrus}$\uparrow$ & {Spectralis}$\uparrow$ & {Topcon}$\uparrow$ & {OOD}$\uparrow$ & {IID}$\uparrow$ & {Gap}$\downarrow$ \\
    
    \hline
    
    {U-Net} & 0.477  & 0.492  & 0.632 & 0.534 & 0.769 & 0.235 \\
    

    {U-Net-Lite} & 0.356 & 0.442 & 0.575 & 0.458 & 0.643 & 0.185 \\
    

    {NCA}  & \textbf{0.715}  & \textbf{0.622}  & \textbf{0.759}  & \textbf{0.699} & \textbf{0.829} & \textbf{0.130} \\
    
    \hline
    \end{tabular}%
    
    \label{tab:retouch_results}
\end{table}

For the RETOUCH dataset (table \ref{tab:retouch_results}), demonstrates the capability of the neural cellular automata to perform well in certain tasks. The NCA outperforms both U-Net variants in both IID and OOD performance. However, there are some notable behaviours. The NCA generalises poorly to the Spectralis domain. Likely this is due to the larger visual discrepancy between the Cirrus and Topcon domains to the Spectralis domain. The low average performance of the NCA in the Spectralis domain was caused mostly by a repeated failure to generalize at all to OOD data: some runs would yield accuracies of roughly equivalent (but lower) to the other two domains, but some would fail completely with accuracies as low as 0.10. This indicates some level of instability in the learning process for the NCA. However, this performance degradation is not see in the U-Nets. Both U-Net generalise somewhat adequately to the Topcon domain with low generalisation gaps, but they both fail in the Cirrus and Spectralis domains.


\begin{table}[ht]
    \centering
    \caption{Average out-of-distribution Dice Accuracy for the MNM2 segmentation dataset on each domain (as unseen out-of-distribution data). Also includes the average performance on unseen validation (IID) data and the difference between IID and OOD accuracies.}
    \setlength{\tabcolsep}{6pt}
    \begin{tabular}{r|ccc|ccc}
    \hline

    & {} & {} & {} & {Mean} & {Mean} & {Mean} \\
     & {GE}$\uparrow$ & {Philips}$\uparrow$ & {Siemens}$\uparrow$ & {OOD}$\uparrow$ & {IID}$\uparrow$ & {Gap}$\downarrow$ \\
    
    \hline
    
    {U-Net} & 0.814 & \textbf{0.888}  & 0.575  & 0.759 & \textbf{0.848} & 0.089 \\
    

    {U-Net-Lite} & 0.732 & 0.780 & 0.389 & 0.634 & 0.789 & 0.156 \\
    

    {NCA}  & \textbf{0.836}  & 0.841  & \textbf{0.694}  & \textbf{0.790} & 0.796 & \textbf{0.006} \\
    
    \hline
    \end{tabular}%
    
    \label{tab:mnm_results}
\end{table}

Both the U-Net and the NCA have more consistent performance on the MNM2 dataset with little stability issues as noted on the NCA's instability on RETOUCH's Spectralis domain. By comparison, the notable outlier are the scans taken from the Siemens machines, which have significantly lower accuracy for both the U-Nets and NCA. Overall, the NCA does again generalize better than the U-Net, with an average Dice accuracy improvement of 0.031. Crucially, the IID performance does not degrade significantly in this dataset compared to RETOUCH, with a drop of only 0.006, compared to a drop of 0.130 in RETOUCH. While the U-Net does perform better in IID data, the smaller generalisation gap implies that the NCA is likely better at generalisation overall. With future work it should be possible to improve this baseline performance and thus achieve even stronger generalisation performance as well. It should be noted U-Nets have had an extensive history to improve their baseline performance as well, whereas NCA has not, thus the potential for improvement in NCA is likely quite high. 

\subsection{Discussion \& Future Work}
This work has shown that at their baseline architecture, NCA performs, at worst, barely behind CNNs, and at-best out-performs CNNs by a wide margin. However, there are still areas for improvement with neural cellular automata. Particularly, extensive work has already been done in an attempt to improve the generalization capabilities of U-Nets and other large scale CNN-based models, such as data augmentation, self-supervised learning, and other techniques. Given their success, it is likely that using these for NCA may further improve their robustness to domain shifts, or simply improve their baseline performance on IID data.

Similarly, the NCA has various aspects which could be improved in ways that are not applicable to U-Nets. One issue, was the lack of stability. It may be possible to use techniques from recurrent neural networks to improve stability throughout the inference process. Additionally, more stability improvements during training are likely to address the issues seen on the Spectralis dataset where the NCA failed to generalise at all. Likewise, a result of the stochastic update process is that multiple runs of the same input may result in different answers, given some level of robustness it may be possible to combine these outputs in an ensemble to improve stability and/or accuracy. 
Likewise, exploring different receptive field sizes and filters may also lead to improved performance.

There is also more to explore regarding model architectures and performance. Neural cellular automata are very similar to graph neural networks (a special case of GNNs), as such it is possible that similar robustness benefits can be found in graph neural networks. Vision transformers are becoming increasingly popular in medical image segmentation as well. The self-attention mechanism acts somewhat similarly to the graph-like message passing structure in cellular automata for a single iteration, and so, it is worth comparing their generalization performance in real-world medical scenarios. 

Another facet that was unexplored in this work was the consequence of adding an extra dimension to the input scans. Both the MRI and OCT scans explored in this work are 3-Dimensional natively, but due to computing resource limitations this was reduced to 2 dimensions. Both U-Nets and NCA both have demonstrated capacity for working with more than 2 dimensions, and as such exploring the impact of dimensionality on robustness is required.

\section{Conclusion}

In the beginning of this paper we asked the question ``\textit{Are large models, which are prone to over-fitting on training data, necessary for medical image segmentation?}'', to answer this, we explored neural cellular automata through two medical imaging segmentation tasks. 
By comparing and contrasting two variations of U-Nets, a large scale model well-known in the medical image segmentation field, and a significantly lightweight version, We have shown that traditional U-Net style models do require large parameter counts to perform well - as seen by the failure of the lightweight version to perform adequately. 
However, by utilizing the NCA architecture it is possible to achieve adequate segmentation accuracies with very limited model size. With limited model sizes, the possibility of using these models on a much wider array of devices becomes possible, as devices with low memory and processing speeds may be able to make use of the models.

Additionally, the NCA architecture also comes with greater generalization capacity. While not outperforming the large scale U-Net on validation data from the same distribution as the training data, the NCA outperforms the U-Nets when challenged on data from different distributions. This highlights the potential for further development of NCA in domain generalisation contexts.


\begin{credits}
\subsubsection{\ackname} 
This research was supported by grants from NVIDIA and utilized an NVIDIA Quadro A6000 for running experiments.

\subsubsection{\discintname}
The authors have no competing interests.
\end{credits}
%
%
%
\bibliographystyle{splncs04}
\bibliography{references}

\end{document}